\documentclass[10pt,twocolumn,letterpaper]{article}

\usepackage{cvpr}
\usepackage{times}
\usepackage{epsfig}
\usepackage{graphicx}
\usepackage{amsmath}
\usepackage{amssymb}

\cvprfinalcopy

\usepackage{epstopdf}
\usepackage{epsfig}
\usepackage{bbm}
\usepackage{xcolor}
\def\MT{{\mathcal T}}

\def\etal{{\em et al.\/}\,}

\usepackage[pagebackref=true,breaklinks=true,letterpaper=true,colorlinks,bookmarks=false]{hyperref}

\def\cvprPaperID{****} % *** Enter the CVPR Paper ID here

\ifcvprfinal\fi
\begin{document}

%%%%%%%%% TITLE
\title{Human Pose Estimation with Spatial Contextual Information}

\author{
Hong Zhang$^1$\hspace{1.0cm}Hao Ouyang$^2$\hspace{1.0cm}Shu Liu$^3$\hspace{1.0cm}
Xiaojuan Qi$^5$\\Xiaoyong Shen$^3$
\hspace{1.0cm}Ruigang Yang$^{1,4}$\hspace{1.0cm}Jiaya Jia$^{3,5}$
\vspace{0.2cm}\\
$^1$Baidu Research, Baidu Inc. \quad $^2$Hong Kong University of Science and Technology \\
$^3$YouTu Lab, Tencent  \quad $^4$ University of Kentucky \quad $^5$The Chinese University of Hong Kong
\\
{\tt\small \{fykalviny,ououkenneth,liushuhust,qxj0125,goodshenxy\}@gmail.com} \\
\vspace{-0.02in}
{\tt\small yangruigang@baidu.com \quad leojia@cse.cuhk.edu.hk}}


\maketitle

\begin{abstract}
We explore the importance of spatial contextual information in human pose estimation. Most state-of-the-art pose networks are trained in a multi-stage manner and produce several auxiliary predictions for deep supervision. With this principle, we present two conceptually simple and yet computational efficient modules, namely Cascade Prediction Fusion (CPF) and Pose Graph Neural Network (PGNN), to exploit underlying contextual information. Cascade prediction fusion accumulates prediction maps from previous stages to extract informative signals. The resulting maps also function as a prior to guide prediction at following stages. To promote spatial correlation among joints, our PGNN learns a structured representation of human pose as a graph. Direct message passing between different joints is enabled and spatial relation is captured. These two modules require very limited computational complexity. Experimental results demonstrate that our method consistently outperforms previous methods on MPII and LSP benchmark.
\end{abstract}

\section{Introduction}
\label{sec:intro}
Human pose estimation refers to the problem of determining precise pixel location of important keypoints of human body. It serves as a fundamental tool to solve other high level tasks, such as human action recognition~\cite{wang2013approach,liang2014expressive}, tracking~\cite{cho2013adaptive,xiaohan2015joint} and human-computer interaction~\cite{shotton2013real}. There are already a variety of solutions where remaining challenges include large change in appearance, uncommon body postures and occlusion.

Recent successful human pose estimation methods are based on Convolutional Neural Networks~(CNNs). State-of-the-art methods \cite{hg,cpm,yang2017pyramid} train pose networks in a multi-stage fashion. These networks produce several auxiliary prediction maps. Then the predictions are refined iteratively in different stages until the final result is produced. It needs to learn semantically strong appearance features and prevent gradient vanishing during training.

\begin{figure*}[!bpt]
\begin{center}
   %\fbox{\rule{0pt}{3in} \rule{0.9\linewidth}{0pt}}
  \includegraphics[width=1\linewidth]{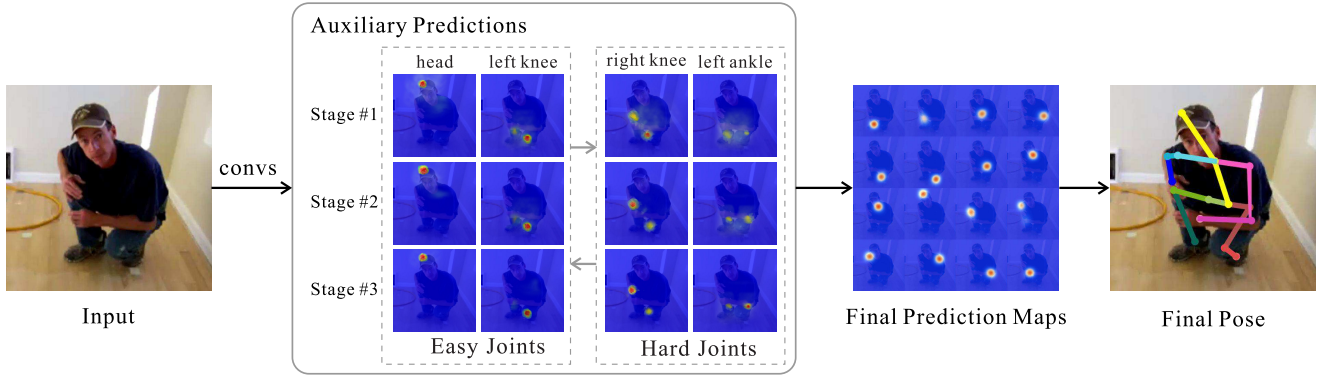}
\end{center}
   \caption{Pipeline of multi-stage prediction. A set of auxiliary predictions are generated. In the first stage, it is easy to identify easy joints while others with severe deformation are still confusing. Relative positions between joints help resolve ambiguity in the second stage. All joints converge to the final prediction in the third stage.}
\label{fig:motivation}
\vspace{-0.1in}
\end{figure*}

Spatial contextual correlation among different joints plays an important role in human pose estimation~\cite{chu2016crf,tompson2015efficient}. Fig.~\ref{fig:motivation} shows the prediction maps of different stages. Rough locations of \emph{head} and \emph{left knee} are easy to identify in the first stage. However, joints like \emph{\{right knee, left ankle\}} in Fig.~\ref{fig:motivation} are with large deformation and occlusion, which are hard to determine only based on the local regions. Fortunately, location of joints like \emph{left knee} is associated with \emph{left ankle}. So the prediction result of \emph{left knee} in the first stage could be indicated as a prior to help infer the location of \emph{left ankle} in the following stage.

Moreover, since human pose estimation is related to structure, it is important to design appropriate guideline to choose directions of information propagation for the joints that are unclear or occluded. Probabilistic Graphical Models (PGMs) are used to facilitate message passing among joints. In \cite{tompson2014joint,chu2016crf}, MRF or CRF is utilized to describe the distribution of human body. Nevertheless, the status of each joint needs to be sequentially updated, which means before updating the status of current joints, status of the previous joints is to be refreshed. The sequential nature of the updating scheme makes it easy to accumulate error. The multilevel compositional models~\cite{zhu2011recursive,tian2012exploring} considered the relations of joints. These methods all rely on hierarchy structures. Pose grammars are based on the prior knowledge of the human body.

To make good use of the underlying spatial contextual information, we propose two conceptually simple and computational efficient modules to estimate body joints.

% In addition, it is important to design the appropriate guideline to choose directions of information propagation for the joints that are unclear or occluded. Probabilistic Graphical Models (PGMs) are used to facilitate the message passing among joints. \cite{tompson2014joint,chu2016crf} utilized MRF or CRF to describe the distribution of human body. Nevertheless, the status of each joint needs to be sequentially updated. Specifically, before updating the status of current joints, it needs to update the status of the previous joints. The sequential nature of the updating scheme makes is accumulate error from the former stages. The multilevel compositional models~\cite{zhu2011recursive,tian2012exploring} are also proposed to consider the relations of joints. A hierarchy structure is designed to capture the pose grammars. However, those methods rely on heuristically designed structures. Pose grammars are based on the prior knowledge of the human body.

\vspace{-0.1in}

\begin{figure*}[!bpt]
\begin{center}
    \includegraphics[width=1\linewidth]{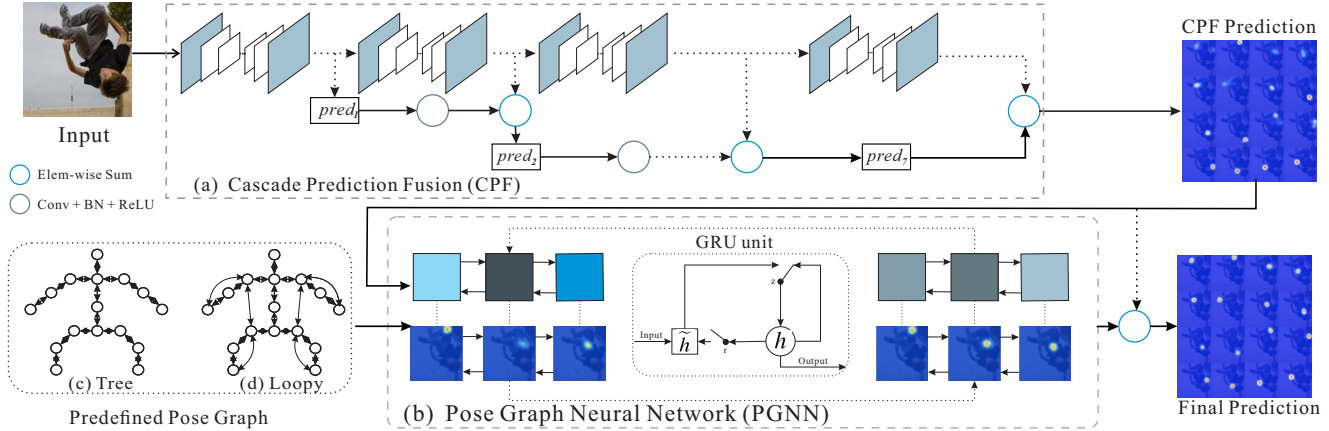}
\end{center}
   \caption{\textbf{Framework.} Our system takes an image as input, and generates the prediction maps. The architecture is with two components where CPF is for computing the prediction maps and the other PGNN is for refining these maps until final prediction.}
\label{fig:overview}
\end{figure*}

\paragraph{\bf Our Contribution \#1}
To utilize the contextual information, we propose \emph{Cascade Prediction Fusion}~(CPF) to make use of auxiliary prediction maps. The prediction maps at previous stage could be deemed as a prior to support predictions in following stages. This procedure is different from that of \cite{hg,cpm}, where prediction maps were concatenated with or added to image feature maps and then fed to following huge CNN trunks. As shown in Fig.~\ref{fig:overview}, we create a light-weight path to gradually accumulate auxiliary prediction maps for final accurate pose estimation. The predictions at different stages are with varied properties. Specifically, predictions from lower layers are with more accurate localization signals while those at higher layers are with stronger semantic information to distinguish among similar keypoints. Our network effectively fuses information from different stages by the shorter path created by CPF.

\vspace{-0.1in}

\paragraph{\bf Our Contribution \#2}
We introduce the \emph{Pose Graph Neural Network}~(PGNN), which is flexible and efficient to learn a structured representation of body joints. Our PGNN is built on a graph that can be integrated in various pose estimation networks. Each node in the graph is associated with neighboring body parts. Spatial relations are thus captured through edge construction. Direct message passing between different nodes is enabled for precise prediction.

Ours is different in modeling spatial relation. PGNN is a novel way to adaptively select the message passing directions in parallel. Instead of defining an explicit sequential order for a human body structure, it dynamically arranges the update sequences. Via simultaneous update, we manage the short- and long-term relation. Finally, PGNN learns a structured graph representation to boost performance.

Our system is also end-to-end trainable, which not only estimates body location but also configures the spatial structures. We evaluate the system on two representative human pose benchmark datasets, \ie, MPII and LSP. It accomplishes new state-of-the-arts with high computational efficiency.

\section{Related Work}
\label{sec:related}
\paragraph{\bf Human Pose Estimation} The key of human pose estimation lies in joint detection and spatial relation configuration. Previous human pose estimation methods can be divided into two groups. The first is to learn feature representation using powerful CNN. These methods detect body joint location directly or predict the score maps for body joints. Early methods like DeepPose~\cite{deeppose} regressed joint locations with multiple stages. Later, Fan~\etal~\cite{fan2015combining} combined local and global features to improve performance. To connect the input and output space, Carreira~\etal~\cite{carreira2016human} iteratively concatenated the input image with previous prediction in each step. Following the paradigm of semantic segmentation~\cite{fcn,deeplab}, methods of~\cite{hg,cpm,yang2017pyramid} used Gaussian peaks to represent part locations. Then a fully convolutional neural network~\cite{fcn} is applied to estimate body joint location. These methods can produce high quality representation and do not predict structure among body joints, however.

The other group focuses on modeling spatial relationship between body joints. The pictorial structures~\cite{pishchulin2013poselet} modeled spatial deformation by designing pairwise terms between different joints. To deal with human poses with large variation, a mixture model is learned for each joint. Yang~\etal~\cite{yang2011articulated} used a part mixture model to infer spatial relation with a tree structure. This structure may not capture very complicated relation. Subsequent methods introduced other models, such as loopy structure~\cite{wang2011learning} and poselet~\cite{pishchulin2013poselet} to further improve the performance.

Later methods~\cite{tompson2015efficient,tompson2014joint} modeled structures via CNN. Tompson~\etal~\cite{tompson2014joint} utilized the Markov Random Field~(MRF) to model distribution of body parts. Convolutional priors were used in defining the pairwise terms of joints. The method of \cite{chu2016structure} utilized geometrical transform kernels to capture relation of joints on feature maps.

\vspace{-0.1in}
\paragraph{\bf Graph Neural Network}
Previous work on feature learning for graph-structure can be divided into two categories. One direction is to apply CNN to graphs. Based on graph Laplacian, methods of \cite{bruna2013spectral,defferrard2016convolutional,kipf2016semi} applied CNN to spectral domain. In order to operate CNN directly on graph, the method of \cite{duvenaud2015convolutional} used a special hash function. The other line focuses on recurrently applying neural networks to each node of the graph. State of each node can be updated based on history and new information passing through edges. The Graph Neural Network~(GNN) was first proposed in \cite{gnn}. It utilized multi-layer perceptrons~(MLP) to learn hidden state of nodes in graphs. However, the contraction map assumption is a restriction. As an extension, the Gated Graph Neural Network~(GGNN)~\cite{ggnn} adopted recurrent gating function~\cite{cho2014learning} to update the hidden state and output sequences. Parameters of the final model can be effectively optimized by the back-propagation through time~(BPTT) algorithm. Very recently, GGNNs was used in image classification~\cite{marino2016more}, situation recognition~\cite{li2017situation} and RGBD semantic segmentation~\cite{qi20173d}.

\section{Our Method}
\label{sec:method}

In this section, we describe the two major components in our method. One is a cascaded multi-stage prediction module where previous-stage prediction serves as a prior to guide present prediction and accumulate auxiliary prediction as shown in Fig.~\ref{fig:overview}(a). The other is to model different parts in a graph, augmented by Pose Graph Neural Network (PGNN) to learn representation, as shown in Fig.~\ref{fig:overview}(b).

\subsection{Cascade Prediction Fusion~(CPF)}
\label{subsec:cpf}

For common pose estimation methods~\cite{hg,cpm}, a set of prediction maps are iteratively refined for body parts. We propose CPF to take the underlying contextual information encoded in auxiliary prediction into consideration.

These prediction maps are in different semantic levels while all of them can be utilized for final predictions. As detailed in \cite{zeiler2014visualizing,chu2017multi}, the lower-layer features focus on local appearance and describe details. It is crucial for accurate joints localization. Meanwhile, the global representations from higher layers help discriminate among different body joints. Our CPF is designed to gradually integrate different semantic information from lower to higher layers. Fig.~\ref{fig:overview}(a) shows the way to incorporate CPF into Hourglass~\cite{hg} framework. It can be built on top of most multi-stage pose estimation frameworks and iterates from the first stage to the final predictions.

For stage $i$, instead of simply fusing the prediction map $pred_{i-1}$ from last stage with $pred_i$ from current stage directly, we provide $pred_{i-1}$ as a prior, which is used for producing $pred_i$. Particularly, the coarse prediction map $pred_{i-1}$ undergoes a $1 \times 1$ convolution to increase channels and is then merged with image features from stage $i$ by using element-wise addition. $pred_i$ is generated by taking the fused feature map as input.

CPF is different from DenseNet~\cite{densenet} and DLA~\cite{dla}. DenseNet emphasizes more on feature reuse and gradient vanish issues. DLA unifies semantic and spatial fusion in the feature level. In contrast, CPF focuses on exploring and aggregating the contextual information encoded in prediction maps.

\subsection{Graph Neural Network~(GNN)}
\label{subsec:gnn}
Graph neural network~(GNN) is a general model handling graph structured data. GNN takes the graph $G = \{K,E\}$ as input where $K$ and $E$ represent the nodes and edges of the graph respectively. Each node $k \in K$ is associated with a hidden state vector $h_k$, which is recurrently updated. The hidden state vector at time step $t$ is denoted as $h_k^t$. The hidden state is updated by taking as input its current state vector and the incoming messages $x_k^t$ from its neighboring nodes $\mathcal{N}_k$. $\mathcal{A}$ is a function to collect messages from neighboring nodes. $\MT$ is a function to update the hidden state. Formally, the hidden state is updated as
\begin{equation}
\begin{aligned}
    x_k^t &= \mathcal{A}(h_u^{t-1}|u \in \mathcal{N}_k), \\
    h_k^{t} &= \MT(h_k^{t-1}, x_k^t).
\end{aligned}
\end{equation}
In the following, we present our new GNN named PGNN for pose estimation.

\paragraph{\bf Graph Construction} Each node $k$ in PGNN represents one body joint and each edge is defined as the connection between neighboring joints. Fig.~\ref{fig:overview}(c) shows an example of how to construct a tree-like graph for human poses. The prediction maps are treated as unary maps learned from a backbone network, which will be detailed in Sec.~\ref{sec:unary}. The hidden state of each node is initialized with its corresponding spatial prediction feature map derived from the original image. The status of node $k$ is initialized as
\begin{equation}
h_k^0 = \mathcal{F}_k(\Theta, I), k \in \{1 \cdots K\},
\label{eq:gnninput}
\end{equation}
where $\mathcal{F}$ indicates the backbone network, $\Theta$ is a set of parameters for the network, and $I$ is the original input image.

\paragraph{\bf Information Propagation} We use the constructed graph to exploit the semantic spatial relation and refine the appearance representation for each joint in steps. Before updating the hidden state of each node, it first aggregates messages of the hidden state at time step $t - 1$ from neighboring node $k'$. As demonstrated in \cite{chu2016structure}, convolutional layers can be used as geometrical transform kernels. It advances message passing between feature maps. It is noted that the weights of convolution for different edges are not shared. So $\mathcal{A}$ is expressed as
\begin{equation}
x_k^t = \sum_{k,k' \in \Omega} W_{p,k} h_{k'}^{t-1} + b_{p,k},
\label{eq:aggregate}
\end{equation}
where $W_{p,k}$ is the convolution weights and $b_{p,k}$ is the bias of the $k^{th}$ node. $\Omega$ is a set of connected edges.

Eq.~(\ref{eq:propagate}) gives the formulation of $\MT$. It updates the $k^{th}$ node with the aggregated messages and the $t-1$ step of hidden state. We follow the same gating mechanism with GRU~\cite{ggnn} and enjoy more computational efficiency and less memory consumption. Again, we utilize convolution operations and do not share weights. $W_{z,k}$, $U_{z,k}$, $b_{z,k}$, $W_{r,k}$, $U_{r,k}$, $b_{r,k}$, $W_{h,k}$, $U_{h,k}$ and $b_{h,k}$ are the weights and biases for the $k^{th}$ node in the update function. With this method, the aggregated information is softly combined with its own memory, which can be expressed as
\begin{equation}
\begin{aligned}
        z_k^t &= \sigma(W_{z, k} x_k^t + U_{z, k} h_k^{t-1} + b_{z, k}), \\
        r_k^t &= \sigma(W_{r,k} x_k^t + U_{r,k} h_k^{t-1} + b_{r, k}), \\
        \tilde{h}_k^t &= tanh(W_{h,k} x_k^t + U_{h,k}(r_k^t \odot h_k^{t-1}) + b_{h,k}),  \\
        h_k^t &= (1 - z_k^t) \odot h_k^{t-1} + z_k^t \odot \tilde{h}_k^t.
        \label{eq:propagate}
\end{aligned}
\end{equation}

\paragraph{\bf Output and Learning} After $T$-time propagation, we get the final prediction
\begin{equation}
\widetilde{P}_k = h^T_k + h^0_k,
\label{eq:finalpred}
\end{equation}
where $h^T_k$ is the final hidden state collected from the corresponding node. $h^0_k$ is the initialization hidden state, which encodes the appearance information of a joint. We get the final prediction by adding these two prediction maps. The graph network is trained by minimizing the $\ell_2$ loss of
\begin{equation}
L_2 = \frac{1}{K}\sum_{k=1}^K\sum_{x,y}||\widetilde{P}_k(x,y) - P_k(x,y)||^2,
\label{eq:loss}
\end{equation}
where $(x,y)$ is the pixel location, $P_k(x,y)$ is the ground truth label at pixel $(x,y)$. $\widetilde{P}_k$ is the prediction map obtained in Eq.~(\ref{eq:finalpred}). The model is trained with back-propagation through time~(BPTT).

\subsubsection{Graph Types}

PGNN can handle a variety of graphs. It develops a novel message passing scheme so that each body receives message from specific neighboring joints. Intuitively, a fully connected graph is expected to be the ideal choice to collect information from all other joints. However, for some joints, such as \textit{head} and \textit{ankle}, it is hard to capture the relationship.

To address this problem, we utilize two types of structure, \ie, tree and loopy structure. It is not known beforehand which one is better. A tree is a simple structure, which captures the relation of neighboring joints. Loopy structure is more complex, allowing message passing in a loop. The structure we use in this paper is illustrated in Fig.~\ref{fig:overview}(c) and (d). Although many tree-like or loopy graphs can be derived, PGNN tackles them in the same way. The graphs are undirected and enable bidirectional message passing.

\subsubsection{Relationship to Other Methods}
Most current state-of-the-art methods focus more on appearance learning of body parts. They capture spatial relation by enlarging the receptive fields. However, poses are with large variation, making the structure information in prediction feature maps still have the potential to boost the performance. Other models like Recurrent Neural Network~(RNN) and Probabilistic Graphical Model~(PGM) can also model the relation. We will detail the difference between PGNN and these models in the following.

\paragraph{PGNN \vs RNN} RNN can be deemed as a special case of PGNN. It is also able to pass information across nodes of a graph, where each body part is denoted as a node and the joints relations are propagated through edges. However, the graph structure requires to be chains for RNN. For its construction, at each time step, the state of current node in RNN is updated by its current state and the hidden state of the parent node. It is different from our PGNN, which collects information from a set of neighboring nodes. Moreover, the order of RNN input is manually defined. A slightly inappropriate setting may destroy the naturally structured relationship of joints.

Tree-structured RNN~\cite{tai2015improved} can handle tree-structured data, which propagates information through a tree sequentially. In addition, before updating the state of subsequent layers $L_t$, it must update the ancestors $L_{t-1}$ at first. Contrarily, PGNN updates all states of the node simultaneously. In addition, RNN shares weights at different time steps. The transfer matrix between nodes in the graph is shared through $T$-time update. Note that in our model, each edge of the graph has different transformation weights.

\paragraph{PGNN \vs PGM} PGNN is also closely related to probabilistic graphical model, which is widely used for pose estimation~\cite{tompson2014joint,chu2016structure} to integrate joint associations. In fact, our model can be viewed as generalization of these models by designing specific update. As detailed in \cite{tompson2014joint}, for a body part $r$, the final marginal likelihood $\widetilde{Q}_r$ is defined as
\begin{equation}
\widetilde{Q}_r = \frac{1}{Z}\prod_{v \in \mathcal{V}}(q_{r|v} * q_v + b_{v \to r}),
\label{eq:mrf}
\end{equation}
where $\mathcal{V}$ is a set of neighboring nodes of $r$. $q_v$ is the joint probability, $q_{r|v}$ is the conditional prior and $b_{v \to r}$ is the bias, respectively. $Z$ is the partition function. When the aggregation function is formulated as the product and update function is represented by Eq.~\eqref{eq:mrf}, PGNN is degraded to the MRF model. With these derivations, it becomes clear that PGNN is a more general model to integrate joint associations by designing specific graph structure and making its own way to update and aggregate functions.

\subsection{Backbone Networks}
\label{sub:backbone}
To verify the generality of our method, we use two backbone networks. One is our modified ResNet-50~\cite{resnet} and the other is the widely used 8-stack Hourglass~\cite{hg}.

\subsubsection{ResNet-50}
\label{subsub:resnet}
ResNet has demonstrated its power on many high-level tasks, including object detection~\cite{resnet} and instance segmentation~\cite{kaiminghe,kaimingliu,liu2018path}. To show the generalization ability, we first modify the ResNet-50 network with a few novel steps for human pose estimation. It achieves decent results, even comparable with using other much deeper networks.

Our strategy is to first convert the vanilla ResNet-50 into a fully convolutional network by removing the final classification and average pooling. We integrate CPF in ResNet-50 and further improve the results using PGNN. The following two techniques are also used to adapt ResNet-50.

\paragraph{Feature Pyramid Network~(FPN)} As introduced in Sec.~\ref{subsec:cpf}, multi-stage prediction is very important for training a pose network. To this end, we adopt Feature Pyramid Network~(FPN)~\cite{fpn} in the vanilla ResNet-50. FPN leverages the pyramid shape in networks for prediction at different feature levels. Similar to FPN, we also use a lateral connection~($1 \times 1$ conv) to merge the information from both bottom-up and top-down pathways. Finally, we produce three auxiliary predictions at three different levels.

\paragraph{Dilated Convolution} Dilated Convolution~\cite{deeplab} is used to enlarge the receptive field without introducing extra parameters. An input image is down-sampled 32 times after fed into the vanilla ResNet-50. However, the feature map is too coarse to precisely localize the joints. To address this problem, we first decrease the stride of convolution layers of the last two blocks from 2px to 1px. This results in shrinking the receptive field. Since for human pose estimation as demonstrated in \cite{hg,cpm}, the spatial information needs to be captured by a large enough area, we replace the $3 \times 3$ convolution layers of the last two blocks with the dilated convolution. Finally, we reduce the stride to 8px.

\paragraph{Other Implementation Details} All models are implemented by Torch~\cite{collobert2011torch7}. We use ImageNet pre-trained model as the base and adopt RMSProp~\cite{tieleman2012lecture} to optimize parameters. The network is trained in a total of 250 epochs with batch size 8. The initial learning rate is $0.001$. It decreased by 10 times at the $200^{th}$ epoch.

\subsubsection{Hourglass}
The 8-stack Hourglass~(Hg) is adopted as the other backbone network to verify our method. It is much deeper than ResNet-50 and is widely adopted by many pose estimation frameworks. With CPF and PGNN integrated in Hourglass, we achieve new state-of-the-art results.

\paragraph{Implementation Details} The network is implemented using Torch and optimized with RMSProp. The parameters are randomly initialized. We train the network in 300 epochs with batch size 6. The learning rate starts at $0.00025$ and decreases by 10 times at the $240^{th}$ epoch.

\section{Experiments}
\label{sec:exp}

\paragraph{\bf Datasets} We evaluate our CPF and PGNN on two representative benchmark datasets, \ie, MPII human pose dataset~(MPII)~\cite{mpii} and extended Leeds Sports Poses~(LSP)~\cite{lsp}. MPII contains about 25,000 images with over 40,000 annotated poses. We use the same setting as that in \cite{tompson2015efficient} to split training and validation sets. The dataset is very challenging since it covers daily human activities with large pose variety. The LSP dataset consists of 11,000 training images and 1,000 testing ones from sport activities.

\paragraph{\bf Data Augmentation} During training, the input image is cropped and warped to size $256 \times 256$ according to the annotated body position and scale. We augment the dataset by scaling it with a factor~($[0.75, 1.25]$), rotation~($\pm 30$), horizontal flipping and illumination adjustment to enhance data diversity, and further improve the robustness of the model for various cases. During testing, we crop the image with the given rough center location and scale of the person. For LSP dataset, we simply use the size and center of the image as rough scale and center.

\paragraph{\bf Unary Maps}
\label{sec:unary}
For the two backbone networks, \ie, ResNet-50 and Hourglass, we take the last prediction score maps as the unary maps. The reason is that the prediction at the final stage is made based on feature with strong semantics, which gathers previous prediction through CPF. It is generally with decent prediction accuracy. The score maps are of size $H \times W \times C$ where $H$ is the height, $W$ is the width, and $C$ is the channel size. In our experiments, the size of $C$ depends on different datasets. $W$ and $H$ are all with size 64, which is $1/8$ of the original input image.

\paragraph{\bf Evaluation Criteria} We use the Percentage Correct Keypoints~(PCK) to evaluate results on the LSP dataset. For MPII, we use PCKh~\cite{mpii}, a modified version of PCK. It normalizes the distance errors with respect to the size of head.

\begin{figure}[!bpt]
\begin{center}
   %\fbox{\rule{0pt}{3in} \rule{0.9\linewidth}{0pt}}
  \includegraphics[width=1\linewidth]{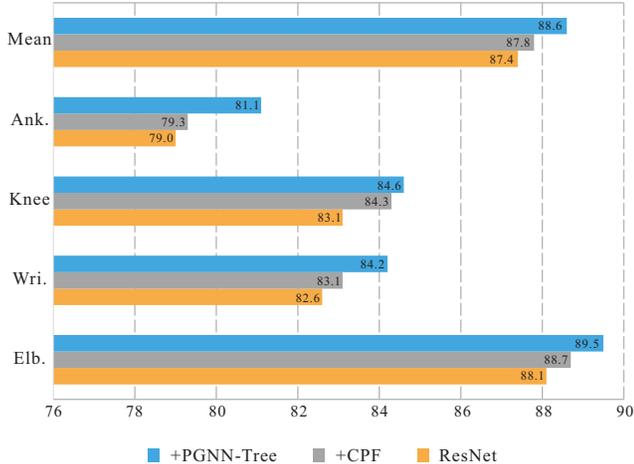}
\end{center}
   \caption{Prediction results of PCKh @0.5 on MPII validation set. We compare the results by adding CPF and further integrating PGNN. The backbone is ResNet-50 and PGNN passes message with a tree-like structure.}
\label{fig:ab_1}
\end{figure}

\subsection{Ablation Study}

To investigate the effectiveness of our proposed CPF and PGNN modules, we conduct ablative analysis on the validation set of MPII Human Pose dataset. We set the modified ResNet-50 as our baseline network. To show the efficacy of our models, all results are tested without flipping or multi-scale testing.

\paragraph{\bf CPF} To evaluate the effectiveness of CPF, we compare results with and without CPF on modified ResNet-50. Fig.~\ref{fig:ab_1} shows results on MPII validation set. ``ResNet" refers to our modified ResNet-50. For ResNet-50 with CPF, some difficult joints like \emph{knee} is $1.2\%$ higher and result of \emph{elbow} is improved by $0.6\%$.

\begin{figure*}[!bpt]
\begin{center}
    \includegraphics[width=1\linewidth]{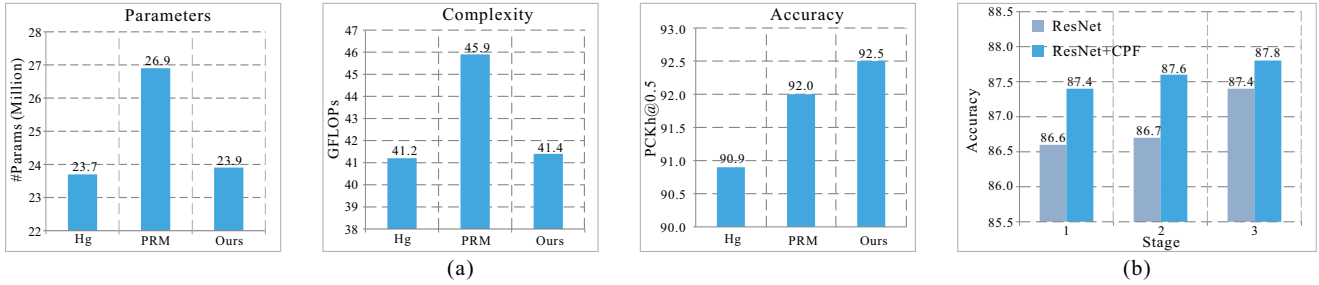}
\end{center}
   \caption{(a) Statistics of parameter numbers, GFLOPs and accuracy on three frameworks, \ie, baseline Hourglass, PRM and our method respectively. (b) Prediction accuracy at different stages with modified ResNet-50.}
\label{fig:complexity}
\end{figure*}

In order to clearly demonstrate accuracy change in each stage. Fig.~\ref{fig:complexity}(b) shows the accuracy produced at different stages. It is observed that the accuracy increases gradually in steps on the ResNet. This manifests that our CPF effectively gathers information from previous predictions.

\begin{figure}[bpt]
\begin{center}
   %\fbox{\rule{0pt}{3in} \rule{0.9\linewidth}{0pt}}
  \includegraphics[width=0.9\linewidth]{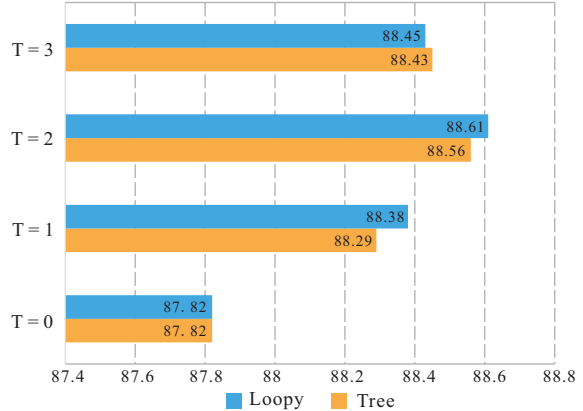}
\end{center}
   \caption{Results at different timesteps with tree-like and loopy-like structure of PCKh @0.5. The backbone is ResNet-50.}
\label{fig:timestep}
\end{figure}
\vspace{-0.1in}

\paragraph{\bf PGNN}
Other than adding CPF, we integrate PGNN to further enhance the accuracy. Fig.~\ref{fig:ab_1} gives the experimental results where ``PGNN" stands for the graph updated twice based on a tree-like structure. The accuracy of parts such as difficult joints of \emph{elbow, wrist} is further improved. The reason is that the contextual information propagated from confident parts through graph helps reduce error.

We use two types of graphs in our experiments. They are tree- and loopy-like graphs in PGNN. Fig.~\ref{fig:timestep} presents the results using different PGNN structures. They are comparable -- connecting the parts including \emph{\{elbow, ankle\}} with other easy parts consistently improves performance. In our experiments, a naive loopy structure, shown in Fig.~\ref{fig:overview}(d), is used. We simply add extra connections, \ie \emph{shoulder}--\emph{wrist}, \emph{ankle}--\emph{hip}, and \emph{shoulder}--\emph{hip}. It is notable that performance of these two types of graphs with the same number of steps are consistent. We thus believe allowing information to propagate between neighboring joints is of great importance. More sophisticated structures may further improve the performance, which will be our future work.

We also conduct experiments to compare results when propagating different times (\ie with varying propagation number $T$) in the system. The results are shown in Fig.~\ref{fig:timestep}. The performance increases by a small amount when increasing $T$, and saturates quickly at $T = 3$. We also notice that propagation is important in the first 2 steps. For the tree-like graph, as revealed in the comparison when applying $T = 0$, $T = 1$ and $T = 2$, we obtain the improvement of around $0.5\%$ and $0.3\%$, respectively. Similar results are observed when using the loopy-like graph. However, the performance begins to drop at $T = 3$. Since it is hard to capture the semantic information between too far away joints, but instead confuses prediction at current joint.

\begin{table}[h]
\setlength{\tabcolsep}{1.5pt}\footnotesize
\begin{center}
\begin{tabular}{c|ccccccc|c}
\hline
Methods~&~Head~&~Sho.&~Elb.~&~Wri.~&~Hip~&~Knee~&~Ank.~&~Mean\\
\hline
Pishchulin~\etal\cite{pishchulin2013poselet} & 74.3  & 49.0  & 40.8  & 34.1  & 36.5  & 34.4 & 35.2 & 44.1 \\
Tompson~\etal\cite{tompson2014joint} & 95.8  & 90.3  & 80.5  & 74.3  & 77.6  & 69.7 & 62.8 & 79.6 \\
Carreira~\etal\cite{carreira2016human} & 95.7  & 91.7  & 81.7  & 72.4  & 82.8  & 73.2 & 66.4 & 81.3 \\
Tompson~\etal\cite{tompson2015efficient} & 96.1  & 91.9  & 83.9  & 77.8  & 80.9  & 72.3 & 64.8 & 82.0 \\
Hu\&Ramanan~\etal\cite{hu2016bottom} & 95.0  & 91.6  & 83.0  & 76.6  & 81.9  & 74.5 & 69.5 & 82.4 \\
Pishchulin~\etal\cite{pishchulin16cvpr} & 94.1  & 90.2  & 83.4  & 77.3  & 82.6  & 75.7 & 68.6 & 82.4 \\
Lifshitz~\etal\cite{lifshitz2016human} & 97.8  & 93.3  & 85.7  & 80.4  & 85.3  & 76.6 & 70.2 & 85.0 \\
Gkioxary~\etal\cite{gkioxari2016chained} & 96.2  & 93.1  & 86.7  & 82.1  & 85.2  & 81.4 & 74.1 & 86.1 \\
Rafi~\etal\cite{rafi2016efficient} & 97.2  & 93.9  & 86.4  & 81.3  & 86.8  & 80.6 & 73.4 & 86.3 \\
Insafutdinov~\etal\cite{insafutdinov16ariv} & 96.8  & 95.2  & 89.3  & 84.4  & 88.4  & 83.4 & 78.0 & 88.5 \\
Wei~\etal\cite{cpm} & 97.8  & 95.0  & 88.7  & 84.0  & 88.4  & 82.8 & 79.4 & 88.5 \\
Chu~\etal\cite{chu2017multi} & 98.5  & 96.3  & 91.9  & 88.1  & 90.6  & 88.0 & 85.0 & 91.5 \\
Chou~\etal\cite{chou2017self} & 98.2  & 96.8  & 92.2  & 88.0  & 91.3  & 89.1 & 84.9 & 91.8 \\
Chen~\etal\cite{chen2017adversarial} & 98.1  & 96.5  & 92.5  & 88.5  & 90.2  & 89.6 & 86.0 & 91.9 \\
Yang~\etal\cite{yang2017pyramid} & 98.5 & 96.7 & 92.5 & 88.7 & 91.1 & 88.6 & 86.0 & 92.0 \\
\hline \hline
% ResNet-50 & 97.8  & 95.8  & 90.2  & 85.4  & 89.8  & 85.5 & 81.2 & 89.8 \\
Newell~\etal\cite{hg} & 98.2  & 96.3  & 91.2  & 87.1  & 90.1  & 87.4 & 83.6 & 90.9 \\
ResNet-ours &　98.2 & 96.4 & 91.6  & 87.1  & 91.2  & 88.0 & 83.6 & 91.2 \\
% Hg-ours & \textbf{98.6}  & 96.8 & 92.6  & 88.6  & 91.4 & 89.3 & 85.8 & 92.2 \\
Hg-ours & \textbf{98.6}  & \textbf{97.0}  & \textbf{92.8}  & \textbf{88.8}  & \textbf{91.7}  & \textbf{89.8} & \textbf{86.6} & \textbf{92.5} \\
\hline
\end{tabular}
\vspace{0.1cm}
\caption{Results of PCKh @0.5 on the MPII test set. Note that ResNet is our modified ResNet-50. ResNet-50 and Hg are all trained with CPF and PGNN.}
\label{tab:mpii}
\end{center}
\end{table}
\vspace{-0.1in}
% (92.2\% when training on full MPII training set)

\subsection{Experimental Results on MPII}

\paragraph{\bf Accuracy} Tab.~\ref{tab:mpii} lists our results on MPII test set. ``Hg-ours" is trained on MPII combined with LSP. The results are produced with five-scale input with horizontal flip testing. Our method trained based on Hourglass yields result $92.5\%$ PCKh at threshold 0.5, which is the highest on this dataset at the time of paper submission. For the challenging parts such as \emph{knee} and \emph{ankle}, we obtain improvement of $2.4\%$ and $3.0\%$ compared to the baseline Hourglass, respectively. Particularly, our method outperforms the method \cite{chu2017multi} with CRF as well. It is noteworthy that accuracy of our method~(ResNet-ours) is also higher than the baseline ResNet-50, which proves the generalization ability.

\paragraph{\bf Complexity} In Fig.~\ref{fig:complexity}(a), we compare the number of parameters and computational complexity between Hourglass, previous method PRM~\cite{yang2017pyramid} and our model. We note that the PRM adds 13.5\% extra parameters compared with Hourglass, while our model only increases parameters by 0.8\%. Additionally, we introduce very limited computation overhead (measured by GFLOPs) on Hourglass, contrary to much increased computational cost from PRM. Our results are with higher quality and decent computational efficiency.

\begin{figure}[bpt]
\begin{center}
   %\fbox{\rule{0pt}{3in} \rule{0.9\linewidth}{0pt}}
  \includegraphics[width=0.9\linewidth]{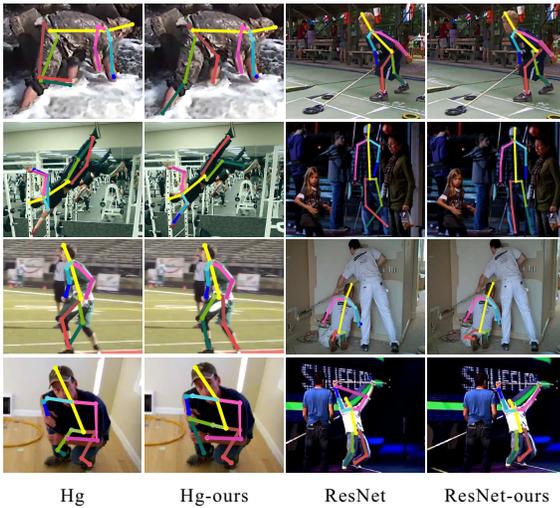}
\end{center}
   \caption{Results on MPII test set produced by different backbone networks, \ie Hourglass and ResNet-50. Hg-ours and ResNet-50-ours are both trained with CPF and PGNN.}
\label{fig:compare_case}
\end{figure}
\vspace{-0.1in}

\begin{figure*}[bpt]
\begin{center}
   %\fbox{\rule{0pt}{3in} \rule{0.9\linewidth}{0pt}}
  \includegraphics[width=0.9\linewidth]{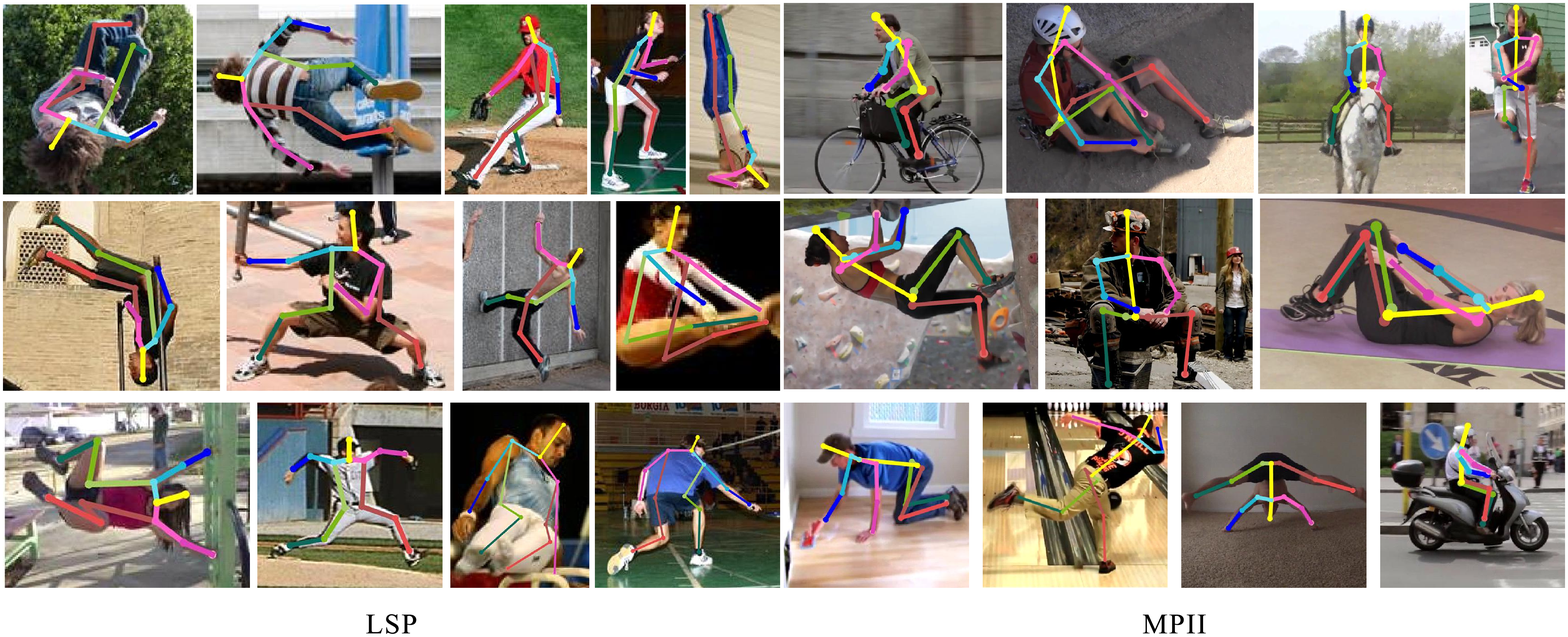}
\end{center}
   \caption{Example output on the LSP and MPII test data.}
\vspace{-0.1in}\label{fig:case_more}
\end{figure*}

\paragraph{\bf Visual Analysis} In Fig.~\ref{fig:compare_case}, we visualize results of baseline and our model. The baseline model has difficulty in distinguishing among symmetric parts and uncommon body postures. For example, in \emph{\{col.1, row.2\}} of Fig.~\ref{fig:compare_case}, \emph{ankle} with large deformation is hard to identify with inherent ambiguity. Our proposed CPF and PGNN provide an effective way to utilize contextual information to reduce the confusion. As a result, the associated joints \emph{knee} help inferring the precise location of \emph{ankle} in our model as shown in \emph{\{col.2 and row.2\}} of Fig~\ref{fig:compare_case}. More results on MPII and LSP generated by our method are shown in Fig.~\ref{fig:case_more}.

\subsection{Experimental Results on LSP} Tab.~\ref{tab:lsp} gives comparison with person-centric annotation. The results are evaluated with PCK scores at threshold 0.2. Following previous methods~\cite{chu2017multi,yang2017pyramid}, we add the MPII training set to the extended LSP training set. Our modified ResNet-50 outperforms most of the methods trained with deeper networks.

\begin{table}[bpt]
\setlength{\tabcolsep}{1.3pt}\footnotesize
\begin{center}
\begin{tabular}{c|ccccccc|c}
\hline
Methods~&~Head~&~Sho.&~Elb.~&~Wri.~&~Hip~&~Knee~&~Ank.~&~Mean\\
\hline
Belagiannis\&Zisserman \cite{belagiannis2017recurrent} & 95.2 & 89.0 & 81.5 & 77.0 & 83.7 & 87.0 & 82.8 & 85.2 \\
Lifshitz~\etal\cite{lifshitz2016human} & 96.8 & 89.0 & 82.7 & 79.1 & 90.9 & 86.0 & 82.5 & 86.7 \\
Pishchulin~\etal\cite{pishchulin16cvpr} & 97.0 & 91.0 & 83.8 & 78.1 & 91.0 & 86.7 & 82.0 & 87.1 \\
Insafutdinov~\etal\cite{insafutdinov16ariv} & 97.4 & 92.7 & 87.5 & 84.4 & 91.5 & 89.9 & 87.2 & 90.1 \\
Wei~\etal\cite{cpm} & 97.8 & 92.5 & 87.0 & 83.9 & 91.5 & 90.8 & 89.9 & 90.5 \\
Bulat\&Tzimiropoulos \cite{bulat2016human} & 97.2 & 92.1 & 88.1 & 85.2 & 92.2 & 91.4 & 88.7 & 90.7 \\
Yang~\etal\cite{yang2017pyramid} & 98.3 & 94.5 & \textbf{92.2} & 88.9 & \textbf{94.4} & \textbf{95.0} & 93.7 & 93.9 \\
\hline \hline
ResNet-ours & \textbf{98.5} & 94.0 & 89.9 & 86.9 & 92.3 & 93.5 & 92.7 & 92.5 \\
Hg-ours & 98.4 & \textbf{94.8} & 92.0 & \textbf{89.4} & \textbf{94.4} & 94.8 & \textbf{93.8} & \textbf{94.0} \\
\hline
\end{tabular}
\vspace{0.1cm}
\caption{Comparison of PCK @0.2 on the LSP dataset. ResNet is short for ResNet-50. Both backbones are trained with CPF and PGNN.}
\label{tab:lsp}
\end{center}
\end{table}

\section{Concluding Remarks}
We have presented effective {\it Cascade Prediction Fusion} (CPF) and {\it Pose Graph Neural Network} (PGNN) to explore contextual information for human pose estimation. CPF makes use of rich contextual information encoded in the auxiliary score maps to produce enhanced prediction. PGNN, differently, is adopted to provide an explicit information propagation scheme to refine prediction. These two components are independent while beneficial to each other in human pose estimation. They are also general for most existing pose estimation networks to boost performance. Our future work will be to extend our framework to 3D and video data for deeper understanding of the temporal and spatial relationship.
%\thispagestyle{empty}

%%%%%%%%% ABSTRACT
% \input{abs}
% %%%%%%%%% BODY TEXT
% \input{intro}
% \input{related}
% \input{method}
% \input{backbone}
% \input{exp}
% \input{conclusion}

{\small
\bibliographystyle{ieee}
\bibliography{egbib}
}

\end{document}